\documentclass{article}
\usepackage{ijcai24}
\usepackage{times}
\usepackage{soul}
\usepackage{url}
\usepackage[hidelinks]{hyperref}
\usepackage[utf8]{inputenc}
\usepackage[small]{caption}
\usepackage{graphicx}
\usepackage{amsthm}
\usepackage{amsmath,amssymb,amsfonts}
\usepackage{booktabs}
\usepackage{algorithm}
\usepackage{algorithmic}

\usepackage{multirow}
\usepackage{tabularx}
\usepackage{enumitem}
\usepackage{bbm}
\usepackage{xcolor}
\usepackage{balance}
\usepackage[flushleft]{threeparttable}

\usepackage{color, colortbl}
\definecolor{Gray}{gray}{0.9}

\title{Graph Neural Networks for Brain Graph Learning: A Survey}

\author{
Xuexiong Luo$^1$\and
Jia Wu$^1$\and
Jian Yang$^1$\and
Shan Xue$^1$\and
Amin Beheshti$^1$\and \\
Quan Z. Sheng$^1$\and
David McAlpine$^2$\and
Paul Sowman$^{3,4}$\and
Alexis Giral$^5$\And
Philip S. Yu$^6$
\affiliations
$^1$School of Computing, Macquarie University, Sydney, Australia\\
$^2$Department of Linguistics, Macquarie University, Sydney, Australia\\
$^3$School of Psychological Sciences, Macquarie University, Sydney, Australia\\
$^4$School of Clinical Sciences, Auckland University of Technology, Auckland, New Zealand\\
$^5$Systemethix, Australia \\
$^6$Department of Computer Science, University of Illinois Chicago, USA
\emails
xuexiong.luo@hdr.mq.edu.au,
\{jia.wu, jian.yang, emma.xue, amin.beheshti, michael.sheng, david.mcalpine and paul.sowman\}@mq.edu.au,
alexis.giral@systemethix.com.au, psyu@uic.edu
}

\begin{document}
\maketitle
\begin{abstract}
Exploring the complex structure of the human brain is crucial for understanding its functionality and diagnosing brain disorders. Thanks to advancements in neuroimaging technology, a novel approach has emerged that involves modeling the human brain as a graph-structured pattern, with different brain regions represented as nodes and the functional relationships among these regions as edges. Moreover, graph neural networks (GNNs) have demonstrated a significant advantage in mining graph-structured data. Developing GNNs to learn brain graph representations for brain disorder analysis has recently gained increasing attention. However, there is a lack of systematic survey work summarizing current research methods in this domain. In this paper, we aim to bridge this gap by reviewing brain graph learning works that utilize GNNs. We first introduce the process of brain graph modeling based on common neuroimaging data. Subsequently, we systematically categorize current works based on the type of brain graph generated and the targeted research problems. To make this research accessible to a broader range of interested researchers, we provide an overview of representative methods and commonly used datasets, along with their implementation sources. Finally, we present our insights on future research directions. The repository of this survey is available at \url{https://github.com/XuexiongLuoMQ/Awesome-Brain-Graph-Learning-with-GNNs}.
\end{abstract}

\section{Introduction}
Analyzing brain structure remains a challenging research problem, yet it is critically important for exploring brain functions and diagnosing disorders. Advancements in neuroimaging technology have enhanced our understanding of brain neuroscience and significantly improved diagnostic capabilities. For instance, functional magnetic resonance imaging (fMRI) primarily uncovers the functional activities of different brain regions of interest (ROIs) by assessing changes in blood oxygenation level dependent (BOLD) signals. Diffusion tensor imaging (DTI) analyzes the structural connectivity among ROIs based on the density of white matter fibers. Electroencephalography (EEG) records the variations in electrical waves during brain activity. Although intelligent diagnosis methods, such as lesion detection \cite{asad2023computer}, have been proposed to employ these neuroimaging data for disorder analysis directly, they often fall short in effectively extracting useful structural information about the human brain for deeper functional mechanism analysis. To address this, a novel data analysis method using graph-structured patterns has emerged, which models the human brain as a brain graph (or brain network) based on neuroimaging data, where different brain regions are represented as nodes and the functional relationships among brain regions as edges. However, traditional brain graph analysis methods \cite{sporns2018graph} tend to focus more on evaluating statistical characteristics through graph theory, such as degree profiles, node centralities, and so forth. These methods suffer from low efficiency and cannot achieve intelligent diagnosis. 

\begin{figure*}[ht!]
		\centerline{\includegraphics[width=1\linewidth]{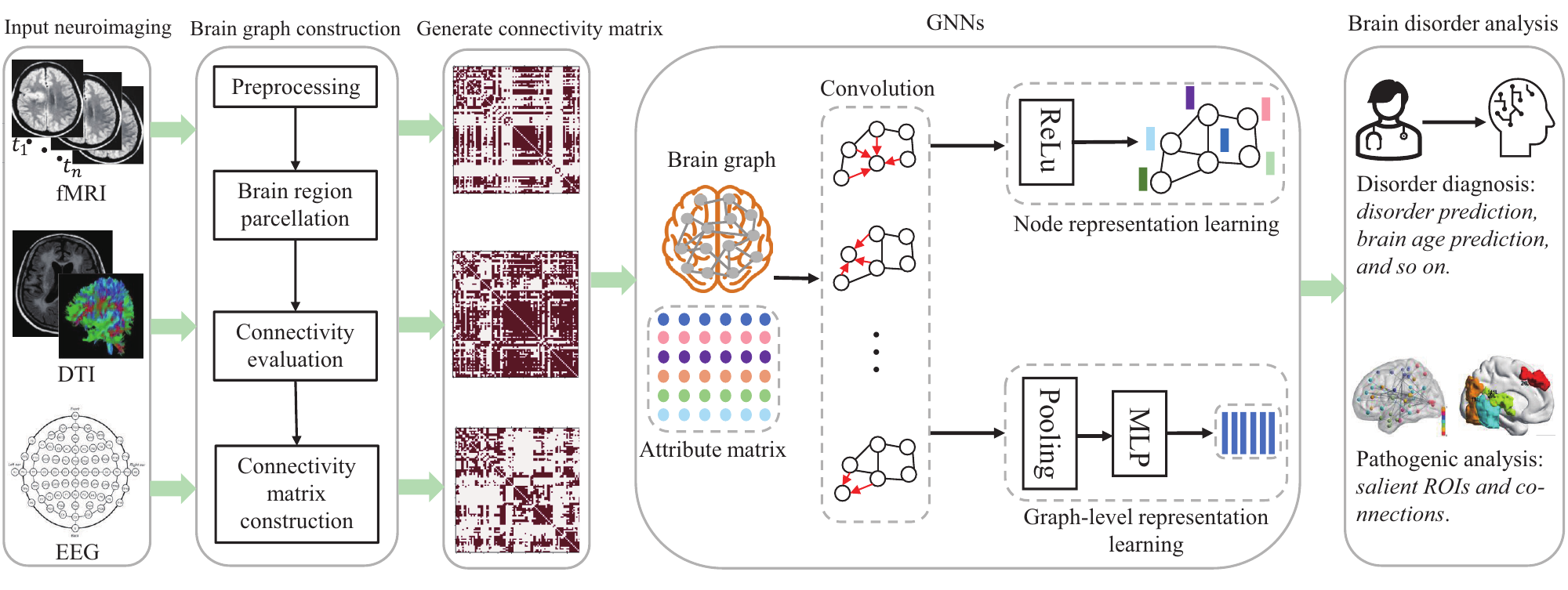}}
        \centering
		\caption{An illustration of brain graph learning framework with GNNs, where different types of neuroimaging are preprocessed to generate corresponding connectivity matrixes. Then, brain graphs are fed into GNNs to learn brain graph representations for disorder analysis.}\label{fig1}
\end{figure*}

GNNs \cite{wu2020comprehensive} have made substantial progress in graph-structured data mining recently. GNNs are designed to learn node representations by aggregating features from the node itself and its neighbors, facilitating various downstream tasks, such as node classification \cite{zhou2019meta}, and link prediction \cite{tan2023bring}. Additionally, GNNs have been successfully applied in real-world scenarios, like drug discovery \cite{bongini2021molecular} and molecular prediction \cite{guo2021few}. Given these advancements, applying GNNs for brain graph mining has become an increasingly studied area. Current research primarily focuses on modeling various brain graphs using neuroimaging data and then designing corresponding GNN models to learn representations of these brain graphs. Furthermore, these studies aim to improve brain disorder prediction, essentially brain graph classification, and pathogenic analysis. The latter involves identifying salient ROIs and connections pertinent to specific disorders, as illustrated in Figure \ref{fig1}.

While some existing articles have endeavored to review the field's progress to date in brain graph analysis, there has been little effort to systematically summarize these studies specifically from the perspective of brain graph learning with GNNs. With this paper, we aim to bridge this gap. We first introduce foundational knowledge of graphs and GNNs. Subsequently, we detail the process of brain graph modeling based on common neuroimaging data, such as fMRI, DTI, and EEG, focusing on two main research problems: \emph{disorder prediction} and \emph{pathogenic analysis}. Next, we propose a systematic taxonomy of existing methods in brain graph learning. Specifically, we categorize these works into three groups based on the type of brain graph generated: static brain graphs, dynamic brain graphs, and multi-modal brain graphs. For each category, we introduce representative approaches tailored to different research problems. Additionally, we offer an overview of libraries with implementations of these methods and commonly used datasets. Finally, we discuss future research directions to advance brain graph learning. The primary goal of this paper is to represent a novel perspective for brain disorder analysis and encourage more researchers to contribute to the burgeoning field of brain graph mining.   
\section{Preliminaries}
This section begins by providing the basic definitions of a graph. Following this, we introduce a unified formulation of GNNs. Next, we detail the process of constructing brain graphs based on three types of neuroimaging data. Finally, we define the two main research problems that are currently prevalent in the field of brain graph learning.
\subsection{Graph Definitions}
A given graph $\mathcal{G}$ generally consists of the following elements: $\left\{V, E, \mathbf{A}, \mathbf{X}\right\}$, where $V=\left\{v_{1},v_{2},\cdot\cdot\cdot, v_{N}\right\}$ and $E=\left\{e_{ij}, i,j \in V \right\}$. These are the node and edge sets, respectively. $\mathbf{A} \in \mathbb{R}^{N\times N}$ is the adjacency matrix, where $\mathbf{A}_{ij}=1$ if an edge exists between node $v_{i}$ and $v_{j}$, otherwise, $\mathbf{A}_{ij}=0$. $\mathbf{X}=\left\{\mathbf{x}_{1},\mathbf{x}_{2},\cdot \cdot \cdot, \mathbf{x}_{N}\right\}$ is the attribute feature matrix, where $\mathbf{x}_{i} \in \mathbb{R}^{d}$ represents attribute information of node $v_{i}$ with $d$ denoting the feature dimensions. Thus, the generated brain graphs based on neuroimaging also include the elements above.
\subsection{Graph Neural Networks}
Currently, a wide array of GNN methods for processing graph-structured data have emerged \cite{wu2020comprehensive}. These methods primarily share a common underlying principle, being the \emph{message-passing mechanism}. This mechanism recursively aggregates features from both the node itself and its neighbors to learn node representations. Consequently, a unified formulation for the convolution operation of GNNs can be expressed as follows:
\begin{equation}
	\mathbf{h}_{v}^{(l+1)}=COM\left ( \mathbf{h}_{v}^{(l)},\left[AGG \left( \left\{\mathbf{h}_{u}^{(l)} \, | \, \forall u \in \mathcal{N}_{v}\right\} \right) \right] \right),
\end{equation}
where $\mathbf{h}_{v}^{(l)}$ is the node representations at the $l$-th layer. $\mathcal{N}_{v}$ denotes the set of neighbors of node $v$, and $\mathbf{h}_{v}^{(0)}$ is initialized using the node attribute feature $\mathbf{x}_{v}$. $COM(\cdot)$ and $AGG(\cdot)$ respectively represent the combination and aggregation functions in the GNN. Here, the output of the GNN is node-level representations of the graph suitable for node-level tasks, such as node classification and link prediction. Notably, graph pooling methods have been extensively explored to obtain graph-level representations of the graph for graph classification and graph generation tasks. The basic operation of graph pooling is as follows:
\begin{equation}
        \mathbf{h}_{G}^{(l)}=READOUT\left\{\mathbf{h}_{v}^{(l)} \, | \, \forall v \in V\right\},
\end{equation}   
where $\mathbf{h}_{G}^{(l)}$ are the graph-level representations in the $l$-th layer, and $READOUT(\cdot)$ is the graph pooling function, such as mean, sum, and max-pooling.
\subsection{Brain Graph Construction}
In current brain graph learning research, three common types of neuroimaging data are predominantly used to construct brain graphs. We introduce vital steps of brain graph construction for these neuroimaging data.

\textbf{fMRI-Driven Brain Graphs.} The fMRI data for one given subject first needs to be preprocessed \cite{cui2022braingb}, which generally involves format conversion, removing the start data, slice timing correction, head motion correction, normalization, smoothing, and temporal filtering. Then, a brain atlas template, like the automated anatomical labeling (AAL) atlas, is used to compartmentalize the brain into different ROIs of a specific number depending on the selected atlas. Next, all ROIs' corresponding BOLD signal series are extracted, and Pearson correlation coefficients for all ROI pairs are computed to construct a functional connectivity matrix. Finally, the functional matrix is generated into the adjacency matrix of the brain graph via the k-nearest neighbour (KNN) method or the threshold setting method. In general, existing methods consider the functional matrix or the adjacency matrix to be an attribute feature matrix, which conducts the graph convolution operation for brain graph learning \cite{cui2022braingb}.

\textbf{DTI-Driven Brain Graphs.} DTI data must be preprocessed as a first step \cite{cui2022braingb}. This will involve format conversion, distortion removal, head motion correction, brain extraction, co-registration, reconstructing local diffusion patterns, and tractography. The brain is then parcellated into different ROIs following the selected atlas and the number of fibers in each ROI is calculated. The structure connectivity between ROIs is constructed according to the number of fibers. Similarly, the structure connectivity matrix generates the adjacency matrix and attribute feature matrix.

\textbf{EEG-Driven Brain Graphs.} EEG data must also be preprocessed \cite{klepl2022eeg}. This will include removing related power-noise frequencies, frequency downsampling, data correction, segment division, and creating frequency bands. The electrodes of the EEG become the nodes of the brain graph, and corresponding connectivity measurement methods are used to construct the edges based on their recorded signals to generate a connectivity matrix. Similarly, the connectivity matrix generates the adjacency matrix and attribute feature matrix.

\subsection{Research Problems}
As an emerging research perspective in brain disorder analysis, brain graph learning is significant in addressing specific research problems. By reviewing existing works on brain graph learning that employ GNNs, we identify two critical research problems that are the primary focus of these studies:

\textbf{Problem 1.} \emph{\textbf{Disorder Prediction.} Disorder prediction aims to utilize GNNs to learn brain graph-level representations and then predict brain graph labels, i.e., diagnose whether the subject suffers from the given disorder.} 

\textbf{Problem 2.} \emph{\textbf{Pathogenic Analysis.} Pathogenic analysis aims to recognize important nodes or edges of brain graphs for the final prediction in the brain graph learning process, i.e., detect which brain regions or connections are related to the brain disorder.}
\begin{figure}[ht!]
		\centerline{\includegraphics[width=1\linewidth]{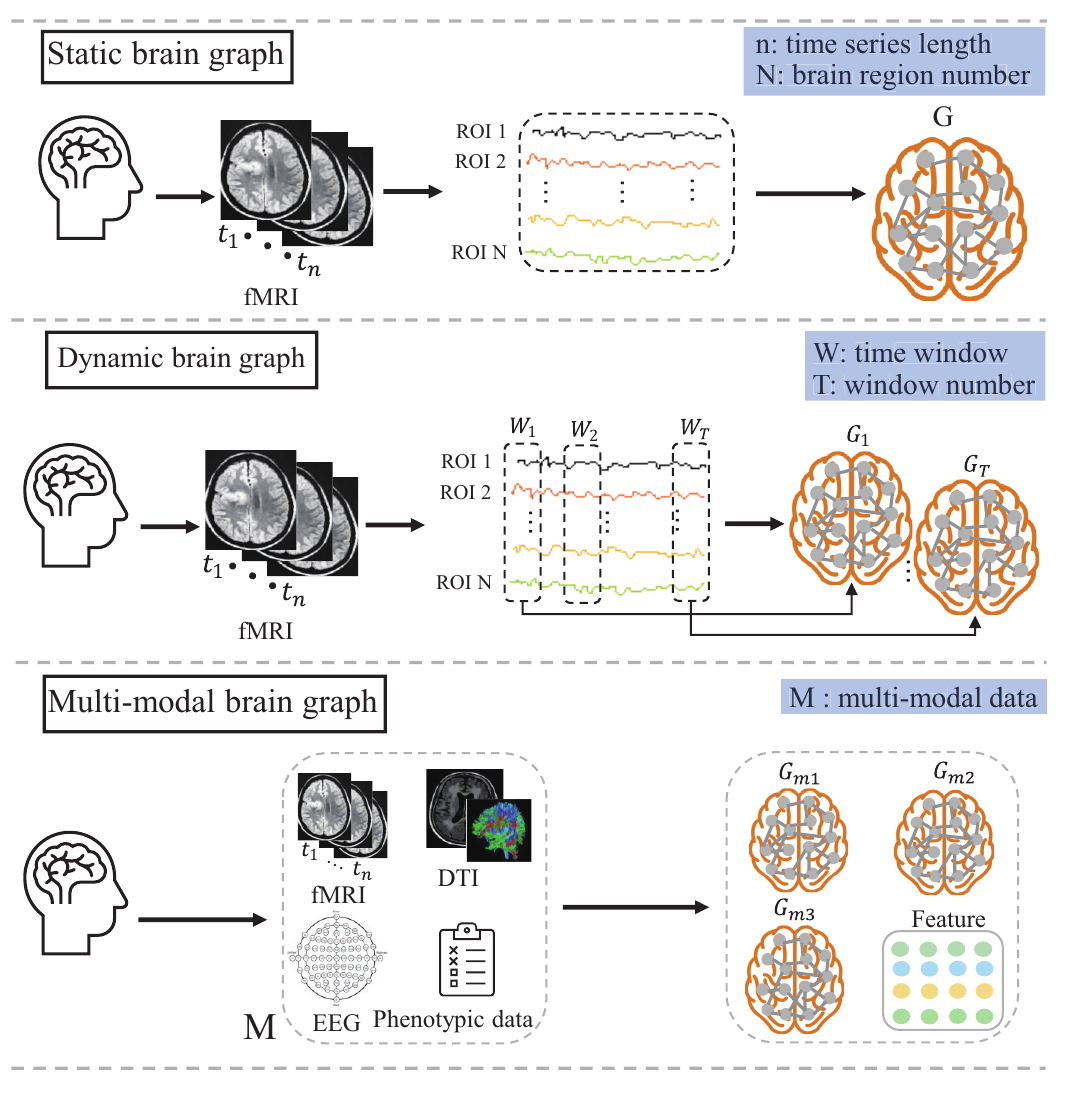}}
        \centering
		\caption{A toy example of different types of brain graph generated.}\label{fig2}
\end{figure}
\section{Taxonomy of Brain Graph Learning Approaches}
In this survey, we consider three types of brain graphs commonly used in brain graph learning, as illustrated in Figure \ref{fig2} with a toy example. Within each brain graph type, the methods are categorized into two groups based on the specific research problem they address. For instance, the prediction-based methods are designed to address Problem 1, while the interpretation-based methods address Problem 2. A library with open-source implementations of the representative methods is provided in Table \ref{tab1}.
\subsection{Static Brain Graph Learning (SBGL)}
Static brain graphs typically represent a single brain graph. While static brain graphs are commonly derived from DTI data, they also involve fMRI and EEG data. For example, fMRI data comprise a series of scanned slices captured over a period of time. Some methods circumvent the temporal aspect by integrating all slices for a subject into a singular, static brain graph. In this section, we explore existing works on SBGL, dividing them into two parts: \emph{prediction-based methods} and \emph{interpretation-based methods}.

\textbf{Prediction-Based Methods.} Typically, the aim of prediction-based SBGL is to design a GNN-specific model that will learn brain graph-level representations and perform disorder prediction, i.e., brain graph classification, to judge whether a subject suffers from a given disorder. For example, BRAINNETTF \cite{kan2022brain} designs a graph transformer-based model for brain graph learning, where the connection profiles of the brain nodes are the attribute features instead of using the positional features of a Transformer model. BRAINNETTF involves a graph pooling method with soft clustering of the brain's functional modules and orthonormal projection to obtain graph-level representations for disorder prediction. To capture the hierarchical structure of ROIs for higher-level representations, THC \cite{dai2023transformer} utilizes a \emph{Transformer-based encoder} with pairwise attention and a clustering layer. This framework learns a globally shared clustering assignment and integrates clustering embedding of multiple layers to learn hierarchical feature representations. Conversely, Com-BrainTF \cite{bannadabhavi2023community} preserves community information in a transformer encoding of a brain graph. 

Several approaches have focused on optimizing the convolution process within brain graph learning to achieve superior brain graph representations. For example, BN-GNN \cite{zhao2022deep} introduces reinforcement learning to perform the convolutional operation adaptively. hi-GCN \cite{jiang2020hi} relies on a hierarchical framework with a two-level GCN \cite{kipf2016semi} designed for a single brain graph and a population brain graph. Furthermore, recent studies also consider addressing the problem of limited disorder data, largely because GNN-based brain graph learning models easily suffer from overfitting and poor generalization ability on insufficient training data. Thus, MML \cite{yang2022data} and PTGB \cite{yang2023ptgb} both offer a multi-task meta-learning method for pre-training a GNN-based brain graph model for later fine-tuning on the target task data for the idea being to improve disorder prediction. 

\textbf{Interpretation-Based Methods.} In essence, interpretation-based brain graph learning aims to provide interpretable analysis results to support disorder prediction, i.e., to recognize the biomarkers associated with disorder-specific brain regions and connections. For instance, using the community-clustering method, tfMRI \cite{li2019graph} first divides a brain graph into subgraphs. Importance scores for the subgraphs are then evaluated based on a salience mapping method. Once the model classifier reaches optimal performance, the biomarkers associated with the specific brain disorder, i.e., the important brain regions, are revealed. Similarly, another work \cite{luokdbrain} introduces the knowledge distillation to extract important brain subgraphs for analyzing the pathogenic reason. To incorporate the prior information to identify salient subnetworks, MVS-GCN \cite{wen2022mvs} constructs multi-view brain graphs with different thresholds and highlights important connections guided by the functional subnetworks.

Furthermore, some methods design the interpretable graph pooling layer to highlight important node features for pathogenic analysis. For example, BrainGNN \cite{li2021braingnn} proposes an R-pool module to keep informative nodes at each graph pooling process and highlight salient ROIs. Other similar methods have also been developed, such as LG-GNN \cite{zhang2022classification}, GAT-LI \cite{hu2021gat} using explainable GNN methods, and IGS \cite{li2023interpretable} with graph sparsification pooling based on important edge selection. We also observe that numerous methods based on generating explanations have also been developed. These approaches first generate a new brain graph and then evaluate important information in the brain graph to give analysis results. IBGNN \cite{cui2022interpretable} generates a globally shared mask across all subjects and evaluates all the edge weights plus the weight sums of all nodes to find important brain regions and connections related to specific disorders. FBNETGEN \cite{kan2022fbnetgen} first conducts feature reduction and denoising. Then, it generates individual brain graphs based on processed time-series features to identify the difference between the given graph and standard brain graphs. However, these methods still generate a shared brain graph to highlight salient ROIs, and they cannot recognize biomarkers within a single brain graph. Hence, to find an important local structure within a single brain graph, BrainIB \cite{liu2023towards} uses the information bottleneck to generate an interpretable brain subgraph toward the corresponding brain graph label.

\subsection{Dynamic Brain Graph Leaning (DBGL)}
Given that fMRI data encapsulates multiple scans of brain activity over a period, capturing the spatio-temporal information in brain graphs is pivotal for unveiling intricate details of the brain's structures. Contemporary studies have leveraged time series-based fMRI data to construct dynamic brain graphs by setting time windows. In essence, this approach results in a subject being represented by a series of brain graphs, each corresponding to a different time window. Accordingly, we have summarized existing studies on dynamic brain graph learning in the following two categories.

\textbf{Prediction-Based Methods.} Prediction-based DBGL mainly involves building spatio-temporal GNNs from brain graph data to learn a representation at the graph level. For example, Multi-Head GAGNN \cite{yan2022modeling} models dynamic brain graphs instead of a single brain graph, to capture spatial and temporal patterns from the given fMRI data. This method first uses a graph U-Net model with multi-head attention to capture spatial patterns. Then the obtained spatial features guide a multi-head attention network learning for the temporal pattern extraction. This approach proves to be beneficial when trying to evaluate the state of an individual brain graph. To accomplish other prediction tasks, ST-GCN \cite{gadgil2020spatio} utilizes a subsequence of time series-based fMRI data to construct dynamic brain graphs. It then applies the spatio-temporal brain graph convolution to help predict gender and age. However, both methods rely solely on an atlas to generate the brain graphs and so the graphs are of a fixed size, which ignores multi-scale structure information. Recently, MDGL \cite{ma2023multi} uses two different atlases to partition the brain into different-sized brain graphs. Hence, this approach does capture multi-scale structure information. Dynamic brain graphs for each scale structure are then constructed, after which a graph isomorphism network captures the spatial patterns and a Transformer network captures the temporal patterns. This multi-scale information is then fused together for disorder prediction.   
\begin{table*}[ht!]
\centering
% \scriptsize
\resizebox{1\textwidth}{!}{
\begin{tabular} {c|c|c|c|c|c}
\toprule
\textbf{Methods} & \textbf{Data Modalities} & \textbf{Research Problems} & \textbf{Datasets} & \textbf{Venue} & \textbf{Code Links} \\
\midrule
\rowcolor{Gray}
\multicolumn{6}{c}{Static Brain Graph Learning}\\
\midrule
tfMRI$^{[1]}$ & fMRI & Disorder Prediction, Pathogenic Analysis &  Biopoint & MICCAI'19 & \url{-}\\
\midrule
hi-GCN$^{[2]}$ & fMRI& Disorder Prediction & ABIDE, ADNI  & Comput. Biol. Med.'20 & \url{https://github.com/hao jiang1/hi-GCN}\\
\midrule
BrainGNN$^{[3]}$ & fMRI & Disorder Prediction, Pathogenic Analysis &  Biopoint, HCP & Med. Image Anal.'21 & \url{https://github.com/xxlya/BrainGNN_Pytorch}\\
\midrule
BRAINNETTF$^{[4]}$ & fMRI & Disorder Prediction & ABIDE, ABCD & NeurIPS'22 & \url{https://github.com/Wayfear/BrainNetworkTransformer} \\
\midrule
BN-GNN$^{[5]}$ & fMRI, DTI, EEG & Disorder Prediction & HIV, BP, ADHD, HA, HI, GD & Neural Netw.'22 &  \url{https://github.com/RingBDStack/BNGNN} \\
\midrule
MML$^{[6]}$ & fMRI, DTI & Disorder Prediction & HIV, BP, PPMI  & KDD'22 & \url{https://github.com/Owen-Yang-18/BrainNN-PreTrain} \\
\midrule
MVS-GCN$^{[7]}$ & fMRI & Disorder Prediction, Pathogenic Analysis & ABIDE, ADNI  & Comput. Biol. Med.'22 & \url{https://github.com/GuangqiWen/MVS-GCN} \\
\midrule
LG-GNN$^{[8]}$ & fMRI, PET & Disorder Prediction, Pathogenic Analysis & ABIDE, ADNI & TMI'22 & \url{https://github.com/cnuzh/LG-GNN}\\
\midrule
IBGNN$^{[9]}$ & fMRI, DTI & Disorder Prediction, Pathogenic Analysis & HIV, BP, PPMI & MICCAI'22 & \url{https://github.com/HennyJie/IBGNN.git}\\
\midrule
Com-BrainTF$^{[10]}$ & fMRI & Disorder Prediction &  ABIDE & MICCAI'23 & \url{https://github.com/ubc-tea/Com-BrainTF}\\
\midrule
\rowcolor{Gray}
\multicolumn{6}{c}{Dynamic Brain Graph Learning}\\
\midrule
ST-GCN$^{[11]}$ & fMRI & Disorder Prediction & NCANDA, HCP & MICCAI'20 &  \url{https://github.com/sgadgil6/cnslab_fmri}\\
\midrule
Multi-Head GAGNN$^{[12]}$ & fMRI & Disorder Prediction &  ABIDE, HCP & Med. Image Anal.'22 & \url{https://github.com/JDYan/Multi-Head-GAGNNs}\\
\midrule
MDGL$^{[13]}$ & fMRI & Disorder Prediction & ABIDE & TNSRE'23 & \url{-}\\
\midrule
STpGCN$^{[14]}$ & fMRI & Disorder Prediction, Pathogenic Analysis & HCP & Hum. Brain Mapp.'23 & \url{-}\\
\midrule
\rowcolor{Gray}
\multicolumn{6}{c}{Multi-Modal Brain Graph Learning}\\
\midrule
MVGE-HD$^{[15]}$ & fMRI, DTI & Disorder Prediction & HIV, BP  & ICDM'17 &  \url{-}\\
\toprule
M2E$^{[16]}$ & fMRI, DTI & Disorder Prediction & HIV, BP & AAAI'18 &  \url{https://github.com/yeliu918/M2E}\\
\midrule
MMGL$^{[17]}$ & fMRI, MRI, PET, Phenotypic Data & Disorder Prediction&  TADPOLE, ABIDE  & TMI'20 & \url{https://github.com/SsGood/MMGL}\\
\midrule
Population-GNN$^{[18]}$ & fMRI, Phenotypic Data & Brain Age Prediction & UKB  & ICML'20 &  \url{https://github.com/kamilest/brain-age-gnn.}\\
\toprule
RTGNN$^{[19]}$ & fMRI, DTI & Disorder Prediction & HIV, BP & TKDE'22 &  \url{https://github.com/RingBDStack/RTGNN}\\
\midrule
BrainNN$^{[20]}$ & fMRI, DTI & Disorder Prediction & HIV, BP  & EMBC'22 &  \url{-}\\
\toprule
CroGen$^{[21]}$ & PICo, Hough & Disorder Prediction & PPMI  & BIBM'22 &  \url{https://github.com/GongxuLuo/CroGen}\\
\toprule
Grad-GCN$^{[22]}$ & Structural MRI, PET & Disorder Prediction, Pathogenic Analysis & ADNI & ISBI'22 &  \url{-}\\
\toprule
Cross-GNN$^{[23]}$ & fMRI, DTI & Disorder Prediction, Pathogenic Analysis & ADNI, Xuanxu, PPMI  & TMI'23 &  \url{-}\\
\toprule
\end{tabular}
}
\begin{tablenotes}
\item 
\small{Note: $^{[1]}$\cite{li2019graph}; $^{[2]}$\cite{jiang2020hi}; $^{[3]}$\cite{li2021braingnn}; $^{[4]}$\cite{kan2022brain}; $^{[5]}$\cite{zhao2022deep}; $^{[6]}$\cite{yang2022data}; $^{[7]}$\cite{wen2022mvs};  $^{[8]}$\cite{zhang2022classification};  $^{[9]}$\cite{cui2022interpretable}; $^{[10]}$\cite{bannadabhavi2023community};  $^{[11]}$\cite{gadgil2020spatio};  $^{[12]}$\cite{yan2022modeling};  $^{[13]}$\cite{ma2023multi}; $^{[14]}$\cite{ye2023explainable}; $^{[15]}$\cite{ma2017multi}; $^{[16]}$\cite{liu2018multi}; $^{[17]}$\cite{zheng2022multi};$^{[18]}$\cite{stankeviciute2020population};
$^{[19]}$\cite{zhao2022multi};$^{[20]}$\cite{zhu2022joint};
$^{[21]}$\cite{luo2022multi};$^{[22]}$\cite{zhou2022interpretable};$^{[23]}$\cite{yang2023mapping}.\\
PET: positron emission tomography data; Phenotypic Data: contains cognitive tests, demographic information and biological information; PICo and Hough: Probabilistic Index of Connectivity and Hough Voting are all different whole brain tractography algorithms. 
}
\end{tablenotes}
\caption{A list of representative brain graph learning methods based on GNNs.} 
\label{tab1}
\end{table*}

\textbf{Interpretation-Based Methods.} In general, the goal of modeling dynamic brain graphs is to capture any brain activity changes. To this end, some methods focus on exploring interpretations of the brain's activity for pathogenic analysis. For example, STpGCN \cite{ye2023explainable} first divides fMRI data into multiple frames of equidistant time windows to form a dynamic brain graph. Then, STpGCN proposes a spatio-temporal pyramid graph convolutional network to learn feature representations. The spatio-temporal features are fused via a bottom-up pathway that can decode multiple cognitive tasks. Importantly, STpGCN introduces a model-agnostic interpretation method to highlight important ROIs for each cognitive task. Quantitative analysis experiments are then performed to evaluate the annotated ROIs. Separately, HDGL \cite{jalali2023hdgl} uses a gated recurrent unit to process time series data from the brain regions, generating node feature representations of the brain graph. Spatio-temporal feature information is then captured and a population brain graph is constructed that represents each subject as a node. The edges reflect phenotypic similarity. One of the key ideas behind HDGL is its self-attention pooling, which identifies salient ROIs for explaining brain disorders during the disorder prediction process via node classification task.
\begin{table*}[ht!]
\centering
% \scriptsize
\resizebox{1\textwidth}{!}{
\begin{tabular} {l|c|c|c|c|c}
\toprule
\textbf{Datasets} & \textbf{Data Source} & \textbf{Data Modalities} & \textbf{Brain Graphs} & \textbf{Nodes} & \textbf{Data Links} \\
\midrule
ABIDE$^{[1]}$ & Autism Spectrum Disorder & fMRI & 1,009 & 200 & \url{http://fcon_1000.projects.nitrc.org/indi/abide/} \\
ADNI$^{[2]}$ & Alzheimer Disorder & Structural MRI, PET & 755  & 90 & \url{ http://www.loni.ucla.edu/ADNI/Data/index.shtml.} \\
PPMI$^{[3]}$ & Parkinson Disorder &  DTI &  718 & 84 & \url{https://www.ppmi-info.org/}\\
HIV$^{[3]}$ & Human Immunodeficiency Virus Disorder & fMRI, DTI & 70 & 90 & - \\
BP$^{[3]}$ & Bipolar Disorder& fMRI, DTI & 97 & 82 & -\\
HCP$^{[4]}$ & Human Connectome Project  & fMRI &  3,542 & 268 & \url{https://db.humanconnectome.org/}\\
ABCD$^{[1]}$ & Adolescent Brain Cognitive Development Study & fMRI & 7,901  & 360 & \url{https://abcdstudy.org}\\
ADHD$^{[5]}$ & Attention Deficit Hyperactivity Disorder & fMRI & 83 & 200 & -\\
Biopoint $^{[4]}$ & Autism Spectrum Disorder & fMRI & 118 & 84 & -\\
\midrule
\end{tabular}
}
\begin{tablenotes}
\item 
\small{Note: $^{[1]}$\cite{kan2022brain}; $^{[2]}$\cite{zhou2022interpretable}; $^{[3]}$\cite{yang2022data}; $^{[4]}$\cite{li2021braingnn}; $^{[5]}$ \cite{zhao2022deep}.}
\end{tablenotes}
\caption{A list of commonly used brain graph datasets.} 
\label{tab2}
\end{table*}
\subsection{Multi-Modal Brain Graph Learning (MBGL)}
Multiple neuroimaging, such as fMRI and DTI, are employed simultaneously to assess brain health when diagnosing a brain disorder. Modeling these diverse data into multi-modal brain graphs, MBGL is particularly advantageous for characterizing brain states and uncovering hidden information relevant to specific disorders. However, the challenge lies in effectively fusing multi-modal data to learn brain graph representations.

\textbf{Prediction-Based Methods.} In general, prediction-based MBGL considers multi-modal data processing and fusion to achieve abundant brain graph representations. For example, MMGL \cite{zheng2022multi} uses multi-modal data, including multi-imaging and non-imaging data like demographic information and biological metrics. MMGL first aligns the features in all modalities into a uniform feature dimension. An inter-modal attention matrix learns the modality-shared and modality-specific representations. In this way, MMGL fuses the two representations into modality-aware representations for each subject. The aim is to capture correlation and complementary information from the multi-modal data. The last step is to construct a population brain graph using an adaptive graph structure learning module and jointly optimize the GNNs and structure learning for disorder prediction. However, other methods only consider the complementary information between different modalities, such as M2E \cite{liu2018multi}, RTGNN \cite{zhao2022multi} using tensor decomposition for information fusion, and BrainNN \cite{zhu2022joint} based on multi-view contrastive learning fusion. Some methods introduce a generative model to avoid the problem of missing data in a single modality. For example, CroGen \cite{luo2022multi} designs a translator to map the hidden representations of a given modality to the corresponding missing modalities. The problem is treated as a regression task that involves structure decoding to generate the brain graph. A generative adversarial network then decides whether the graph is real or generated. Differently, other methods aim to construct population brain graphs. topoGAN \cite{bessadok2021brain} clusters population brain graphs constructed from some learned multi-modal graph representations. Multiple brain graphs are then predicted for each cluster by a discriminator. Alternatively, Population-GNN \cite{stankeviciute2020population} introduces a population graph based on neuroimaging and non-imaging modalities to predict the age of the brain. 

In addition, some methods consider capturing community structures to improve graph representations. Specifically, MVGE-HD \cite{ma2017multi} proposes a joint training framework for graph representation learning and hub detection, and designs a weighted fusion strategy for multi-modal data. To alleviate missing attribute features and inject community information into embedding representations, SCP-GCN \cite{liu2019community} defines an fMRI graph as a feature matrix and a DTI graph as a structure matrix. Then, SCP-GCN designs a Siamese framework including two sample inputs to preserve community structures during the graph convolution process.

\textbf{Interpretation-Based Methods.} Interpretation-based MBGL focuses on fusing multi-modal information to yield more reliable and comprehensive analytical results. For example, the work by \cite{zhu2022multimodal} proposes a triplet network to fuse multi-modal brain graphs, where self-attention is used to learn single-modality representations and cross-attention is used to fuse the multi-modal representations. Specifically, the self-attention weight highlights the importance of each brain region, and pathogenic regions are detected by contrasting these important regions with the brain graphs of healthy controls and patients. To improve cross-modal fusion and help detect important ROIs, Cross-GNN \cite{yang2023mapping} generates a dynamic brain graph by defining multi-modal representations as node features and a correspondence matrix as an adjacency matrix. Cross-distillation is then conducted from single-modality and multimodal representations. Finally, the correspondence matrix is used to evaluate disorder-specific biomarkers. However, other methods introduce a feature map to visualize important ROIs for pathogenic analysis after learning multi-modal representations. Grad-GCN \cite{zhou2022interpretable} integrates three modalities of imaging data into a multi-modal brain graph and computes the feature map by gradient class activation mapping to identify important ROIs after the GNN operation.
\section{Brain Graph Datasets}
To advance brain disorder analysis through brain graph learning and encourage broader research participation, we have compiled the commonly used datasets, with their statistics detailed in Table \ref{tab2}. This summary includes each dataset's research source, modality type, number of subjects (i.e., brain graphs), and the number of divided brain regions (i.e., brain graph nodes), thus, offering a comprehensive overview of these resources. Additionally, we have provided links to the public datasets to facilitate easy access.
\section{Future Research Directions}
While there have been significant advancements in brain graph learning with GNNs, the field still faces several unresolved challenges. In this section, we identify and discuss key research directions that merit future exploration. 
\subsection{Reliable Brain Graph Construction} In preprocessing brain neuroimaging data to obtain weight connectivity matrices, the current methods typically either adopt a weight threshold to determine the edges between brain regions or employ a KNN method to generate the structural information. However, setting a weight threshold often involves arbitrary decision-making, making identifying an effectively optimized threshold challenging. Additionally, the absence of explicit brain node features means that using the connectivity matrix as node features in KNN-based graph construction may not always yield an effective brain graph structure. Therefore, designing a reliable method for brain graph construction is crucial, as it significantly impacts the overall quality of brain graph learning. One potential avenue to address these challenges is the application of reinforcement learning to optimize the threshold selection for connectivity matrices tailored explicitly for different brain disorders. Furthermore, employing graph structure learning methods \cite{zhu2021survey} represents another promising research direction for enhancing brain graph structures. 
\subsection{Multi-Scale Brain Graph Fusion} To partition the brain into different brain regions as the definition of brain graph nodes, a corresponding brain atlas needs to be used, such as the commonly used AAL 116, Craddock 200 atlas (CC200), or similar. However, selecting a different brain atlas will generate a different brain graph at the corresponding size. Currently, there are no well-defined criteria for choosing the most suitable brain atlas with which to segment the brain about a specific brain disorder. However, creating multi-scale brain graph structures using multiple atlases for a single subject can be beneficial. This approach helps to depict the connectivity state of brain networks and enriches brain graph representations. While the MDGL method \cite{ma2023multi} constructs multi-scale dynamic brain graphs utilizing both the AAL and CC200 atlases for disorder prediction, it falls short in thoroughly exploring the optimal scale size selection. Additionally, the simplistic approach of concatenating multi-scale features does not substantially enhance the representations of brain graphs. Thus, more research efforts are needed to address the above problems. 
\subsection{Prior Domain Knowledge Fusion} Currently, most research focuses on using neuroimaging data to generate various brain graphs and designing a brain graph-oriented GNN to learn the embedding representations. Yet neuroscientists have reached many conclusions about the brain’s mechanisms. For example, the human brain contains different neural systems, such as the Somato-Motor Network and the Visual Network. These neural systems are often closely related to human behaviours and can affect the occurrence of some brain disorders. We also note that some studies \cite{liu2019community,cui2022interpretable} consider capturing the community structures in brain graphs or performing interpretation analysis from the view of the neural systems. However, these methods do not draw on prior knowledge for their model designs to improve brain graph learning. It could be fruitful to fuse this prior domain knowledge into brain graph learning,  especially disorder interpretation analysis. Doing so may greatly promote research progress in brain disorder analysis.  
\subsection{Brain Subgraph Extraction} As an important component of graph structures, subgraphs have been studied extensively. For example, SubgraphX \cite{yuan2021explainability} uses a Monte Carlo tree search algorithm to identify important subgraphs to explain GNN prediction. SUGAR \cite{sun2021sugar} employs a reinforcement learning method to select important subgraphs and a subgraph neural network that learns graph-level representations. Brain subgraphs play a pivotal role in brain graph learning. On the one hand, identifying discriminative brain subgraphs between healthy individuals and patients can reveal underlying pathogenic factors, as abnormalities in specific brain regions often result in atypical connections with neighboring areas. On the other hand, brain graphs are often characterized by clustering and small-world properties, as evidenced in previous research \cite{bassett2006small}. Extracting brain subgraphs, therefore, is instrumental in exploring pathogenic reasons and enhancing the quality of brain graph representations. Despite its importance, this research area has not received adequate attention and remains underexplored.
\subsection{Brain Graph Augmentation} The collection and preprocessing of neuroimaging data are notably costly, resulting in a paucity of available datasets. This limitation often leads to a small sample size in commonly used datasets, which poses a risk of overfitting and poor generalization ability in deep brain graph models. While some studies have suggested using pre-training strategies or graph contrastive learning methods to mitigate these issues \cite{yang2022data,luo2024constrative}, these approaches still rely on large source datasets for pre-training and require data augmentation for contrastive learning optimization. However, given the subtle differences between healthy and patient brain graphs, traditional graph data augmentation methods, like node dropping and edge perturbation, risk distorting the brain graph structure. Therefore, developing effective brain graph data augmentation techniques remains a critical research area. Drawing inspiration from the successful application of diffusion models in molecular graph generation \cite{xu2022geodiff}, exploring similar methods for brain graph generation presents a promising research direction.  
\subsection{Medical Experimental Evaluation} Current interpretation-based brain graph learning methods, such as those in \cite{cui2022interpretable,luo2024constrative}, primarily assess their interpretative results by comparing identified brain regions and connections with related medical findings. However, this approach relies solely on existing knowledge for validation. What is lacking is a comprehensive loop-locked test and evaluation. Given the ongoing quest to unravel the pathogenic mechanisms of brain disorders, a field still replete with unknowns, these interpretation results from brain graph learning models are not fully corroborated by existing medical findings. Some salient brain regions and connections identified by current models do not align with established medical research conclusions. This discrepancy underscores the need for more rigorous and reliable experimental analyses to validate these interpretations. Therefore, a significant future research direction lies in enhancing the integration of brain graph learning with brain neuroscience research. Specifically, combining medical analysis experiments to evaluate interpretive results is crucial for advancing our understanding and ensuring these computational models' reliability in medical science. 
\section{Conclusion}
In this work, we introduce brain graph construction and current research problems. We systematically review existing studies and discuss future research directions. To promote the development of the research community, we summarize the representative methods and commonly used datasets providing links to the code and data where available. We also provide the repository of this survey and hope this survey will attract more attention and advance brain graph learning for disorder research.     

\section*{Acknowledgments}
This work was supported by the Australian Research Council Projects with Nos. LP210301259 and DP230100899, Macquarie University DataX Research Centre, and NSF under grant III-2106758. J. Wu is the corresponding author. 

\bibliographystyle{named}
\bibliography{ijcai24}

\end{document}